\documentclass{article}

\usepackage{hyperref}

\usepackage{microtype}
\usepackage{graphicx}
\usepackage{subfigure}
\usepackage{booktabs} 
\usepackage{amssymb}
\usepackage{amsmath}
\usepackage{pbox}
\usepackage{caption}
\usepackage{makecell}
\DeclareCaptionLabelFormat{nonestyle}{ }

\usepackage[accepted]{icml2019}

\icmltitlerunning{}

\begin{document}

\twocolumn[
\icmltitle{Group Equivariant Deep Reinforcement Learning}



\icmlsetsymbol{equal}{*}

\begin{icmlauthorlist}
\icmlauthor{Arnab Kumar Mondal}{Mc,Cim}
\icmlauthor{Pratheeksha Nair}{Mc,Bio}
\icmlauthor{Kaleem Siddiqi}{Mc,Cim}

\end{icmlauthorlist}

\icmlaffiliation{Mc}{School of Computer Science, Mcgill University, Montreal, Canada}
\icmlaffiliation{Cim}{Centre for Intelligent Machines, Mcgill University, Montreal, Canada}
\icmlaffiliation{Bio}{McGill Centre for Bioinformatics, McGill University, Montreal, Quebec, Canada}
\icmlcorrespondingauthor{Arnab Mondal}{arnabm@cim.mcgill.ca}

\icmlkeywords{Machine Learning, ICML}

\vskip 0.3in
]



\printAffiliationsAndNotice{}  

\begin{abstract}
In Reinforcement Learning (RL), Convolutional Neural Networks(CNNs) have been successfully applied as function approximators in Deep Q-Learning algorithms, which seek to learn action-value functions and policies in various environments. However, to date, there has been little work on the learning of symmetry-transformation equivariant representations of the input environment state. In this paper, we propose the use of Equivariant CNNs to train RL agents and study their inductive bias for transformation equivariant Q-value approximation. We demonstrate that equivariant architectures can dramatically enhance the performance and sample efficiency of RL agents in a highly symmetric environment while requiring fewer parameters. Additionally, we show that they are robust to changes in the environment caused by affine transformations.
\end{abstract}
\vspace{-0.7cm}
\section{Introduction}

Reinforcement Learning has always faced the challenge of handling high dimensional sensory input, such as that given by vision or speech. To this end, it was demonstrated that a convolutional neural network could directly learn control policies from raw video data, with success in various Atari game environments \cite{mnih2013playing}. More recently, there has been work to improve both the feature extraction from raw images\cite{grattarola2017deep} as well as the underlying Deep Q-Learning algorithm \cite{schaul2015prioritized,horgan2018distributed,van2016deep,wang2015dueling}. Following this, a variety of models focusing on short-term memory \cite{kapturowski2018recurrent}, episodic memory \cite{badia2020never} and meta controlling \cite{badia2020agent57} have been introduced. Despite these advances, the generalization of trained agents to new environments and the improvement of sample efficiency has not been widely explored. One way to tackle this problem is to apply standard regularization techniques such as L2 regularization, dropout \cite{srivastava2014dropout}, data augmentation and batch normalization \cite{ioffe2015batch}, as proposed in \cite{farebrother2018generalization,cobbe2018quantifying}.
Approaches rooted in Meta-RL have also been proposed to address the generalization problem \cite{wang2016learning,dasgupta2019causal,kirsch2019improving}.
\vspace{-0.1cm}

In this work, we exploit the intrinsic properties of an environment, such as its symmetry, to improve the performance of Deep RL algorithms. In particular, we consider the efficacy of using an E(2)-Equivariant CNN \cite{weiler2019general} architecture as a function approximator for training RL agents using an Equivariant Q-Learning algorithm. We show that in a game environment, with a high degree of symmetry, such an approach provides a significant performance gain and improves sample efficiency as it learns from fewer experience samples.
We further show that the inherent inductive bias for the equivariance of symmetry transformation of our proposed approach, enables the effective transfer of knowledge across previously unseen transformations of the environment. Our proposed method is complementary to the other generalization ideas in RL mentioned earlier, and hence can be used in conjunction with them. Using the proposed method adds negligible computational overhead, improves generalization and facilitates a higher degree of parameter sharing. The ideas explored in this paper could be extended to more challenging RL tasks, such as path planning in dynamic environments, where the dynamics is given by symmetry transformation, using aerial views. In such tasks, the policy may be designed to be equivariant to symmetric transformations of the viewpoint.
\vspace{-0.1cm}

 The rest of the paper is organized as follows: Section \ref{bak} gives a brief overview of relevant background.
 In Section \ref{contrib}, we review the theory of E(2)-equivariant convolution and introduce our Equivariant DQN model. Finally, we present empirical results on two environments, Snake and Pacman, in Section \ref{res}, demonstrating the promise of equivariant Deep RL.

\vspace{-0.1cm}

\section{Background}
\label{bak}
Group equivariant CNNs (G-CNN) \cite{cohen2016group} exploit the group of symmetries of input images to reduce sample complexity, learn faster and improve the capacity of CNNs without increasing the number of parameters. This network architecture uses a new convolution layer whose output feature map changes equivariantly with the group action on the input feature map and promotes higher degrees of weight sharing. The theory of steerable CNNs \cite{cohen2016steerable,weiler2019general,weiler2018learning} generalizes this idea to continuous groups and homogeneous spaces. In this work, we focus on using an E(2)-Equivariant Steerable CNN\cite{weiler2019general} architecture for deep RL.
\vspace{-0.1cm}

Given an input signal, CNNs extract a hierarchy of feature maps. The weight-sharing of the convolution layers makes them inherently translation-equivariant so that a translated input signal results in a corresponding translation of the feature maps\cite{cohen2016group}. An E(2)-Equivariant Steerable CNN carries out translation, rotation and reflection equivariant convolution on the image plane. The feature spaces of such Equivariant CNNs are defined as spaces of feature fields and are characterized by a group representation that determines their transformation behaviour under transformations of the input, as discussed in Section \ref{scnn}.


The Deep Q-learning Network (DQN) \cite{mnih2013playing} has been widely used in RL since its inception. The DQN utilizes ``experience replay" \cite{lin1993reinforcement} where the agent's experiences at each time-step are stored in a memory buffer, and the Q-learning updates are done on samples drawn from this buffer, which breaks the correlation between them. A variant of this strategy is the ``prioritized replay buffer" \cite{schaul2015prioritized}, where the experiences are sampled according to their importance. A second variant, the Double DQN or DDQN \cite{van2016deep}, addresses the problem of maximization bias, which occurs due to the usage of the same Q network for the off-policy bootstrapped target.
An additional improvement is the use of an advantage function and the learning of a value function to determine the action-values using a common convolutional feature learning module, in a Dueling Network \cite{wang2015dueling}. We experiment with the above mentioned variants.



\section{Method} \label{contrib}
\subsection{E(2)-equivariant convolution} \label{scnn}

In this section, we briefly describe the theory behind E(2)-equivariant convolution. First, we define the group 
$T(2)\rtimes G$ where $G \leq O(2)$. T(2) is a translational group on $\mathbb{R}^2$ and $G$ is a subgroup of the orthogonal group O(2), which are continuous rotations and reflections under which the origin is invariant. Intuitively, we are dealing with the subgroups of the group of isometries of a 2-D plane called E(2). In contrast to regular CNNs, which work with a stack of multiple channels of features $f:\mathbb{R}^2\to \mathbb{R}$, the steerable CNN defines a steerable feature space of feature fields $f:\mathbb{R}^2\to \mathbb{R}^c$ which associates a $c$ dimensional feature vector $f(x)\in \mathbb{R}^c$ to every $ x \in \mathbb{R}^2$. The feature fields are linked to a transformation law that defines their transformations under the action of a group. The transformation law of a feature field is characterized by the group representation $\rho:G\mapsto \text{GL}(\mathbb{R}^c)$, where $\text{GL}(\mathbb{R}^c)$ represents the group of all invertible $c\times c$ matrices. This defines how each of these $c$ channels mixes when the vector $f(x)$ is transformed. The operator for a transformation $tg$, where $t\in T(2)$ and $g\in G$, is given by:
\begin{equation}
    \left([\text{Ind}_G^{T\left(2\right)\rtimes G}\rho]\left(tg\right).f\right)(x) := \rho(g).f\left(g^{-1}\left(x-t\right)\right)
\end{equation}
where $[Ind_G^{T\left(2\right)\rtimes G}\rho]$ is called the induced representation. Analogous to the channels of a regular CNN, we can stack multiple feature fields $f_i$ with their corresponding representation $\rho_i$ and the stack $\bigoplus_i f_i$ then transforms under $\bigoplus_i \rho_i$, which is a block diagonal matrix. Notice that due to $\rho$ being a block diagonal matrix each feature field transforms independently. Having described the feature fields, we will next give the equation for equivariance and the constraint it imposes on the convolution kernel. Consider two feature fields $f_{in}:\mathbb{R}^2 \to \mathbb{R}^{C_{in}}$ with representation $\rho_{in}$, $f_{out}:\mathbb{R}^2 \to \mathbb{R}^{C_{out}}$ with representation $\rho_{out}$ and a convolution kernal $k:\mathbb{R}^2\to \mathbb{R}^{c_{out}\times c_{in}}$ then the desired equivariance is given by:
 \begin{equation}
 \begin{split}
      k\ast \left([\text{Ind}_G^{T\left(2\right)\rtimes G}\rho_{in}] \left(tg\right).f_{in}\right) = \\
      [\text{Ind}_G^{T\left(2\right)\rtimes G}\rho_{out}]\left(tg\right).(k\ast f_{out})  
\end{split}
 \end{equation}
 where convolution is defined as usual as:
 \begin{equation}
     f_{out}(x) := (k \ast f_{in})(x) = \int_{\mathbb{R}^2}k(y)f_{in}(x+y)dy
 \end{equation}
 This can only be achieved if we restrict ourselves to G-steerable kernels which satisfy the kernel constraint:
\begin{equation}
     k(gx) := \rho_{out}(g)k(x)\rho_{in}(g^{-1}) \quad \forall g\in G \& x\in \mathbb{R}^2
 \end{equation}
 Imposing this constraint on the kernels significantly reduces the number of parameters and promotes parameter sharing. Also, by obtaining equivariance in each convolution layer of the network, they can be composed to extract equivariant features from the input 2D image signal. Further details on the kernel basis are provided in \cite{weiler2019general}.

\subsection{Environment} \label{env}
\begin{figure}[!ht]
	\centering
	\begin{minipage}{.24\columnwidth}
	\captionsetup{labelformat=empty}
		\centering
		\includegraphics[width=\textwidth]{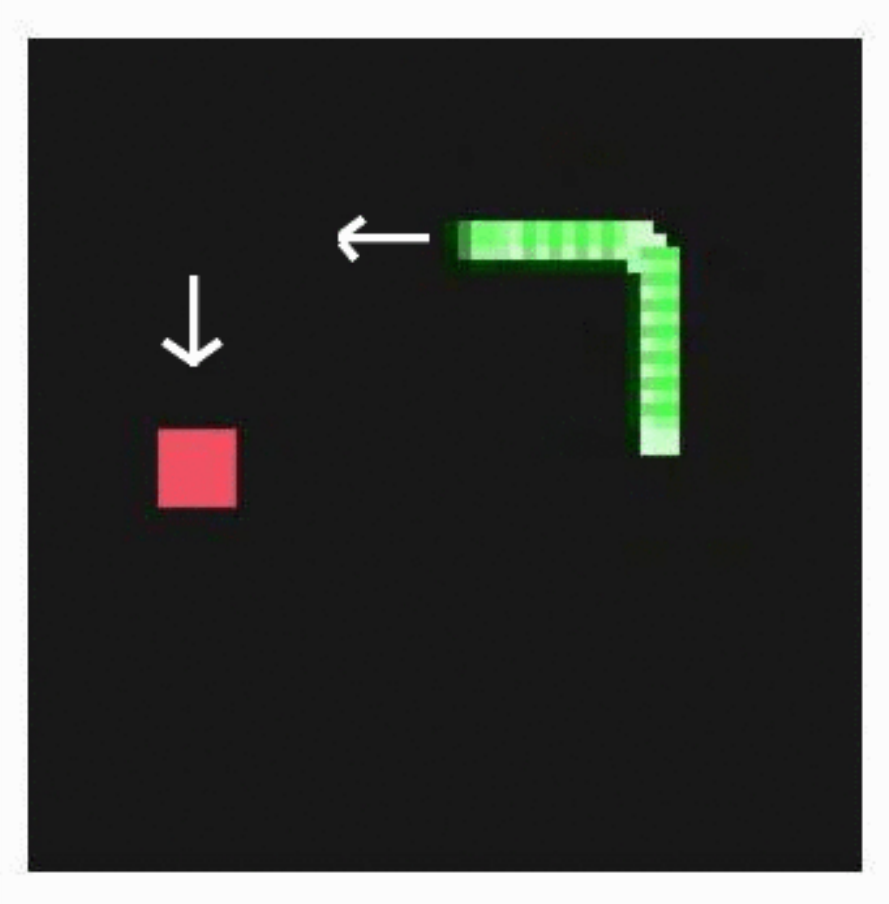}
		\caption*{original ($e$)}
	\end{minipage}%
	\begin{minipage}{.24\columnwidth}
	\captionsetup{labelformat=empty}
		\centering
		\includegraphics[width=\textwidth]{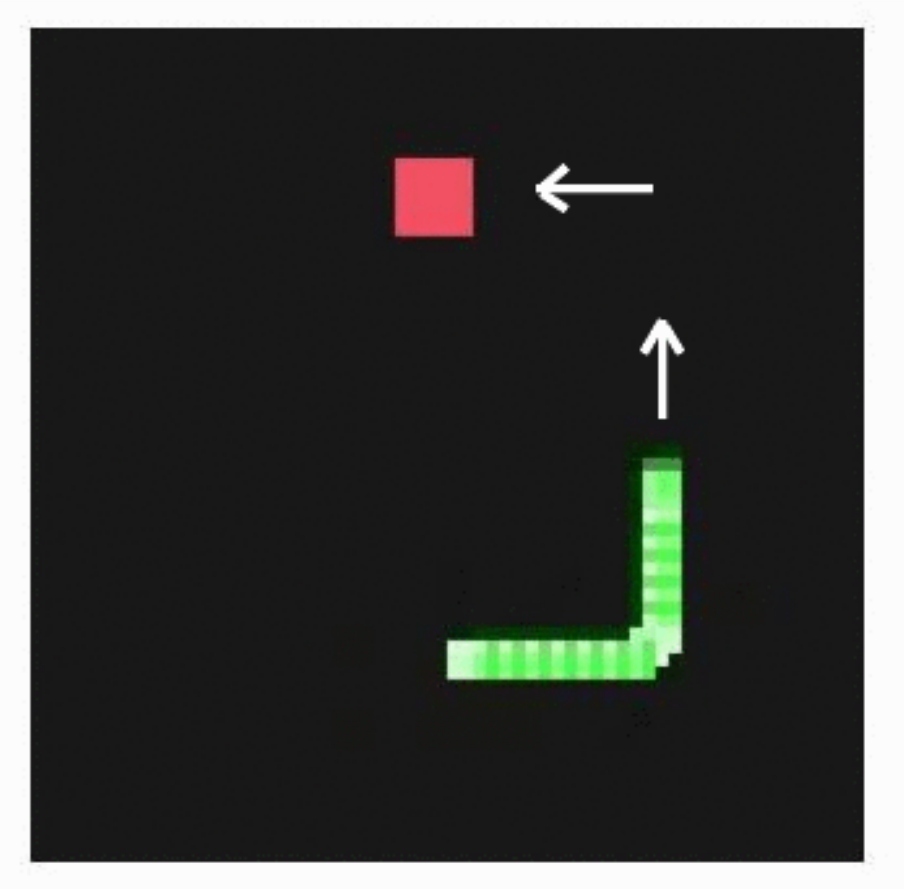}
		  \caption*{$r$}
		\label{label2}
	\end{minipage}
		\begin{minipage}{.24\columnwidth}
		\captionsetup{labelformat=empty}
		\centering
		\includegraphics[width=\textwidth]{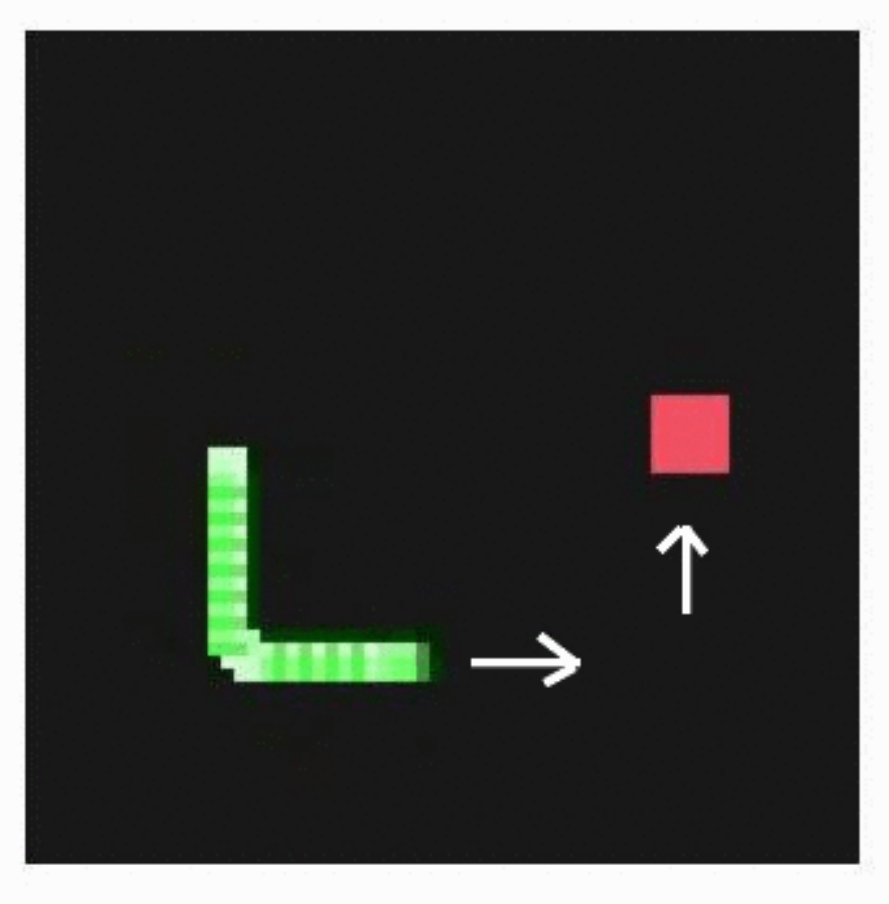}
		\caption*{$r^2$}
		\label{label1}
	\end{minipage}%
	\begin{minipage}{.24\columnwidth}
	\captionsetup{labelformat=empty}
		\centering
		\includegraphics[width=\textwidth]{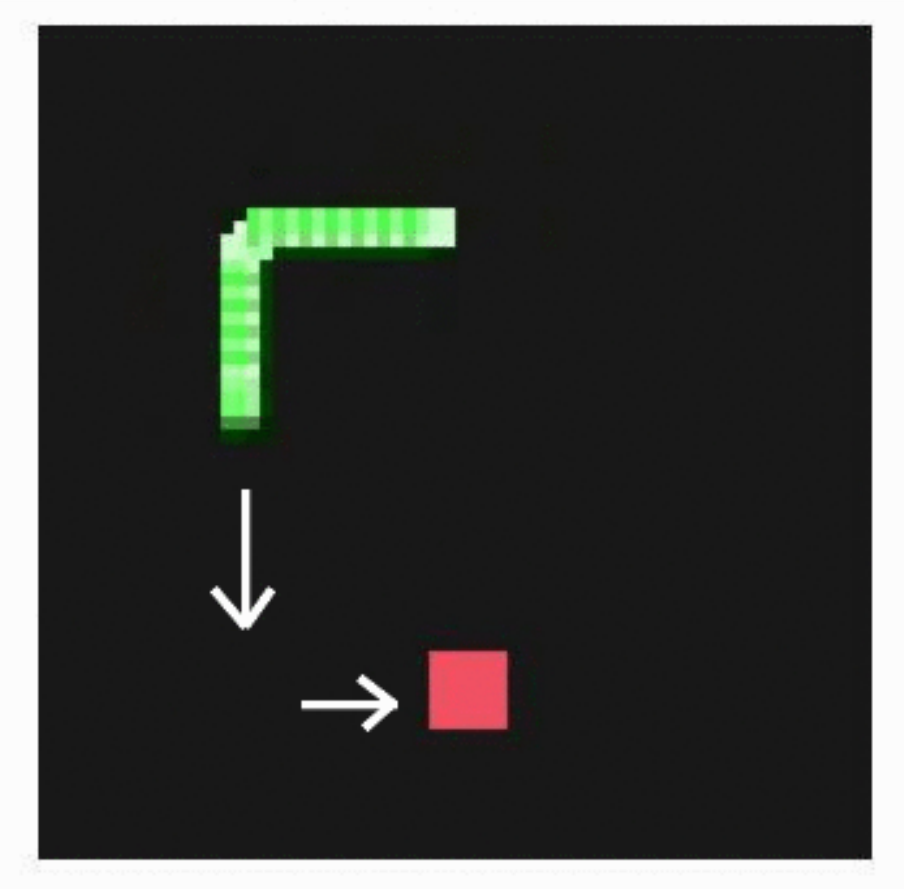}
		\caption*{$r^3$}
		\label{label2}
	\end{minipage}
	
		\begin{minipage}{.24\columnwidth}
		\captionsetup{labelformat=empty}
		\centering
		\includegraphics[width=\textwidth]{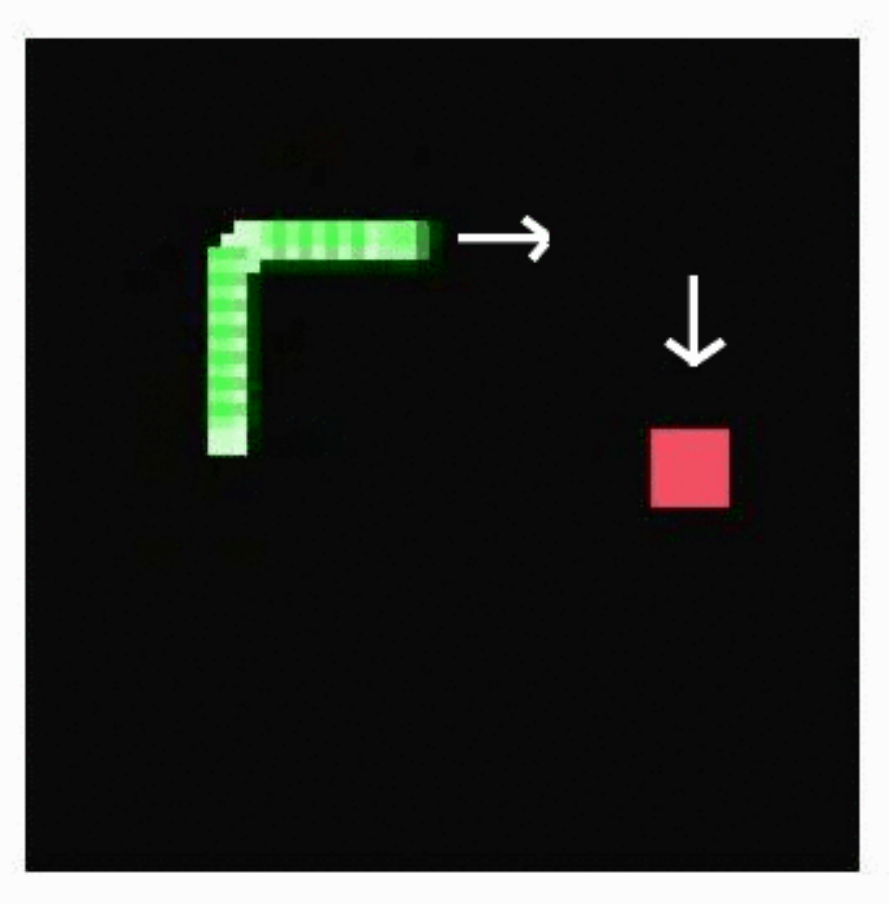}
		\caption*{ $t$}
		\label{label1}
	\end{minipage}%
	\begin{minipage}{.24\columnwidth}
	\captionsetup{labelformat=empty}
		\centering
		\includegraphics[width=\textwidth]{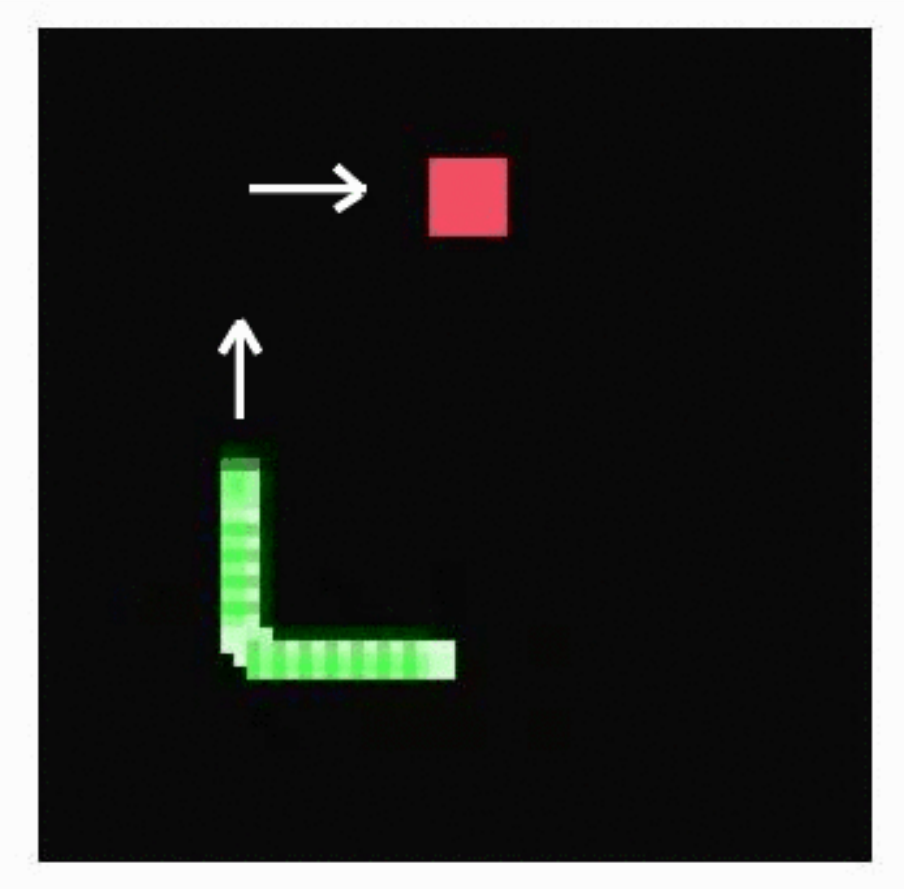}
		\caption*{$tr$}
		\label{label2}
	\end{minipage}
		\begin{minipage}{.24\columnwidth}
		\captionsetup{labelformat=empty}
		\centering
		\includegraphics[width=\textwidth]{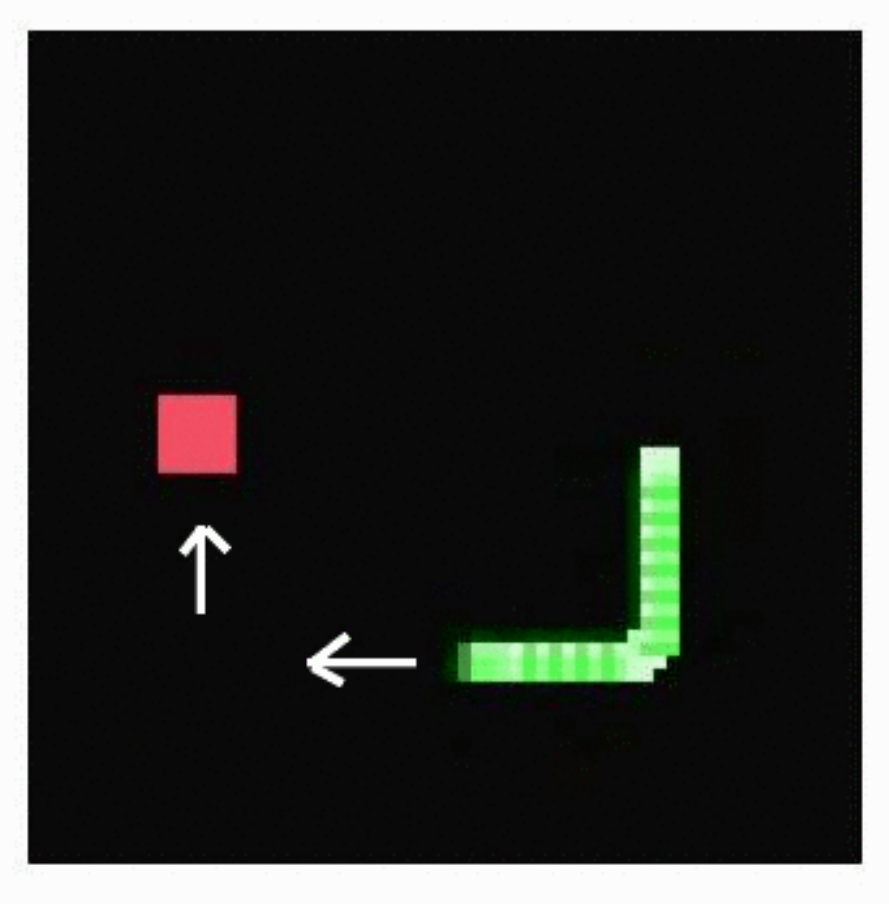}
		\caption*{$tr^2$}
		\label{label1}
	\end{minipage}%
	\begin{minipage}{.24\columnwidth}
	\captionsetup{labelformat=empty}
		\centering
		\includegraphics[width=\textwidth]{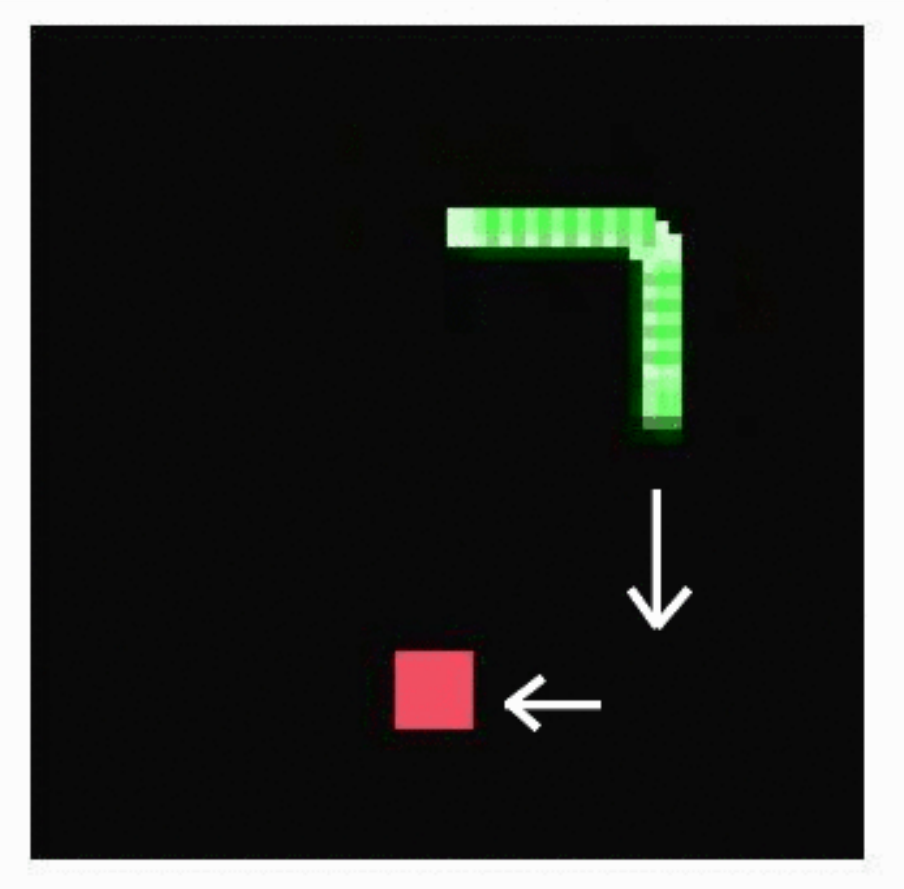}
		\caption*{$tr^3$}
		\label{label2}
	\end{minipage}
	\caption{If we denote the 90 degree clockwise rotation by $r$ and reflection over the vertical axis as $t$, then the group elements of $D_4$ are $\{e, r, r^2, r^3, t, tr, tr^2, tr^3\}$ where $e$ is the identity action. These panels show the action of these group elements (transformations) on a game screen and how they affect the optimal policy (shown by white arrows).}
\label{fig:screen}
\vspace{-0.5cm}
\end{figure}
In this work, we primarily experiment with two environments - the Snake game of the Pygame Learning Environment \cite{tasfi2016pygame} and the Atari Pacman environment \cite{brockman2016openai} \footnote{\href{https://gym.openai.com/envs/MsPacman-v0/}{https://gym.openai.com/envs/MsPacman-v0/}}. 
In the Snake game\footnote{\href{https://pygame-learning-environment.readthedocs.io/en/latest/user/games/snake.html}{https://pygame-learning-environment.readthedocs.io/-\\en/latest/user/games/snake.html}}, the agent is a snake which grows in length each time it feeds on a food particle and gets a reward of +1. The food particle is randomly placed somewhere inside the valid area of a screen. The snake can choose four legal actions: move up, move down, move left, and move right. A terminal state is reached when the snake comes in contact with its body or the walls, and the agent then receives a score of -1. From Figure \ref{fig:screen}, we see that under the action of group elements of $D_4$, the current optimal policy should change equivariantly, which suggests the possible benefits of learning the Q values for each action using equivariant features extracted from the game screen.

The Pacman game consists of a maze, a player agent and a few ghosts. Food particles are placed along the paths of the maze while the ghosts move freely around it. The player agent is also allowed four actions - move up, move down, move right and move left and it gets a positive reward for each particle it consumes without running into any of the ghosts. The game screen has a global $D_1$ symmetry and a degree of local $D_4$ symmetry. 

\subsection{Equivariant Deep Q-Network} \label{model}
\label{implementationdetails}
Henceforth in this paper, ``equivariant convolution" refers to E(2)-equivariant steerable convolution. Suppose our preprocessed input is of dimension $m\times d\times d$ where $m$ is the number of channels, and $d\times d$ is the size of the image. We convert it into a feature field represented by $s=\bigoplus_{i\in I} s_i$ where $I=\{1,..,m\}$ and $s_i$ is an image of dimension $d\times d$. The transformation law of each channel is given by trivial representation ($\rho_{triv}$) of a chosen discrete group ($G$) for each channel. We further choose a regular representation ($\rho_{reg}$) for intermediate feature fields, which are permutation matrices given a group element $g \in G$, to derive the kernel basis of equivariant convolution. Using regular representation preserves the equivariance with point-wise nonlinear activation functions such as ReLU. We stack equivariant convolutions followed by ReLU to obtain an equivariant feature extractor $F_{eqv}:\mathbb{R}^{m\times d\times d}\to \mathbb{R}^{n}$ where $n$ denotes the dimension of extracted equivariant features. The detailed architecture of this feature extractor and its relationship to the vanilla feature extractor we use in DDQN are in Appendix \ref{netarc}. A discussion on how to choose the group and its representation for a feature field along with group restriction is included in Appendix \ref{grprep}. Assuming that we do not restrict the group $G$ along the depth of the network, our transformation rule of the extracted feature vector with respect to the transformation of input is given by:
  \begin{equation}
  \label{eqv}
 \begin{split}
      F_{eqv} \left([\text{Ind}_G^{T\left(2\right)\rtimes G}\rho_{triv}]\left(tg\right) s;\theta \right) = \rho(g)\left(F_{eqv}\left(s;\theta\right)\right)
\end{split}
 \end{equation}
where $\rho(g)=\bigoplus_{j\in J}\rho_{reg}(g)$ and $$\left([\text{Ind}_G^{T\left(2\right)\rtimes G}\rho_{triv}]\left(tg\right)s\right)\left(x\right):= \bigoplus_{i\in I} \left[s_i\left(g^{-1}\left(x-t\right)\right)\right] $$ Note that Equation \ref{eqv} gives the desired equivariance and $J=\{1,..,(n/N)\}$ where $N$ is the order of the $g$. $N$ divides $n$ and $n/N$ is the number of feature fields at the output. Intuitively, Equation \ref{eqv} means that at every feature field the values permute along its dimension when we transform the input by some group element. Also note that if we restrict the group along the depth we will have $g_{res}\in G_{res}\leq G$ in the RHS of Equation \ref{eqv} instead of $g$. Having obtained the feature vector which transforms equivariantly we can add a final linear layer to obtain the Q values:
\begin{equation}
  Q_{eqv}(s,a,\phi) = W_a.F_{eqv}(s;\theta)+b_a  
\end{equation}
where $a\in \mathcal{A}=\{1,..,|\mathcal{A}|\}$, $W_a=[w_{a1} .. w_{an}]$ and $\phi=\{\theta,W_1,..,W_{|\mathcal{A}|},b_1,..,b_{|\mathcal{A}|}\}$(the set of all parameters). The linear layer learns whether or not to preserve the equivariance in output depending on the environment. We use DDQN as our baseline model throughout this work whose final loss at iteration $l$ is given by:
\begin{equation}L(\phi_l) = {\mathbb{E}}_{s,a,r,s'} \left[\left(y_l^{DDQN}-Q_{eqv}(s,a,\phi_l)\right)^2\right] 
\end{equation}
with target:
\begin{equation}
  y_l^{DDQN}= r + \gamma Q_{eqv}(s',\operatorname*{arg\,max}_{a'}Q_{eqv}(s',a',\phi_l),\phi^{-}) 
\end{equation}
where $\phi^{-}$ represents the parameters of the frozen network. The gradients computed through both the linear and the equivariant feature extractor networks are backpropogated to update their parameters.

\section{Experiments} \label{res}
\begin{figure}[ht]
		\begin{minipage}{\columnwidth}
		\centering
		\includegraphics[width=0.95\textwidth]{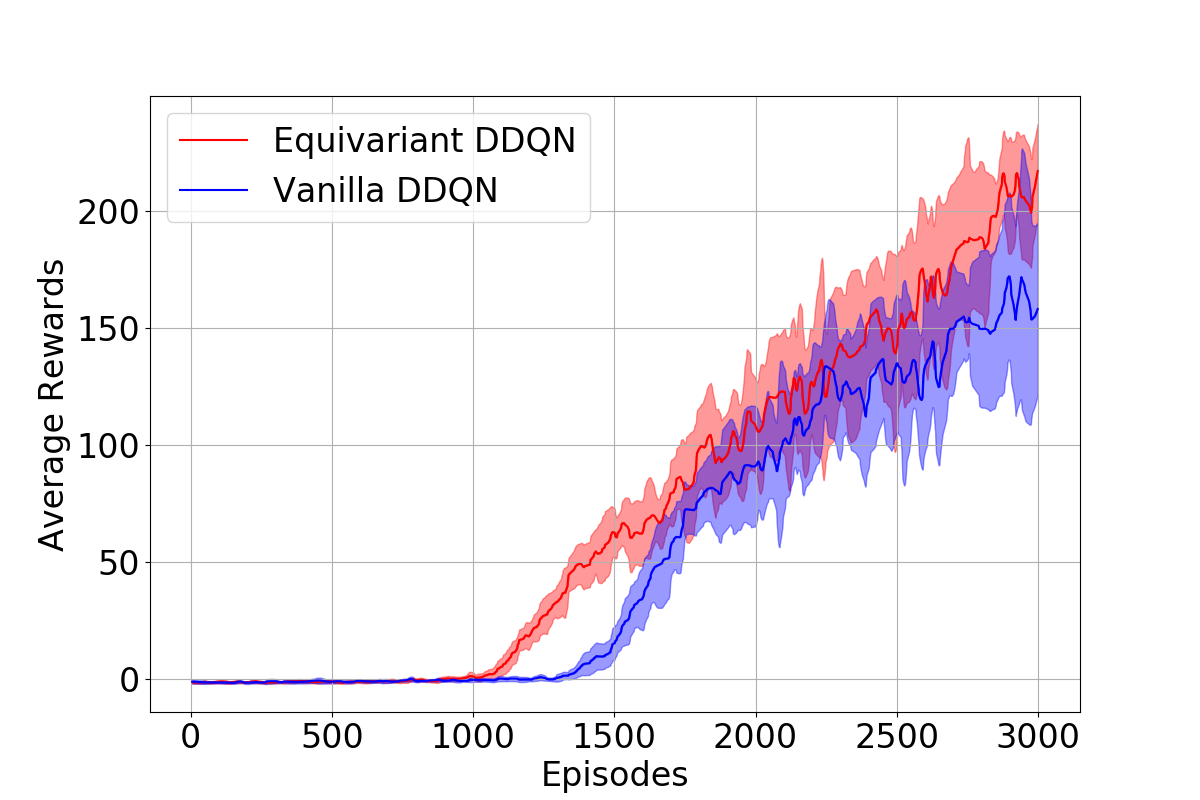}
		\captionsetup{labelformat=empty}
		\caption*{Snake}
		\label{fig:snake}
		\end{minipage}
		\begin{minipage}{\columnwidth}
		\centering
		\includegraphics[width=0.95\textwidth]{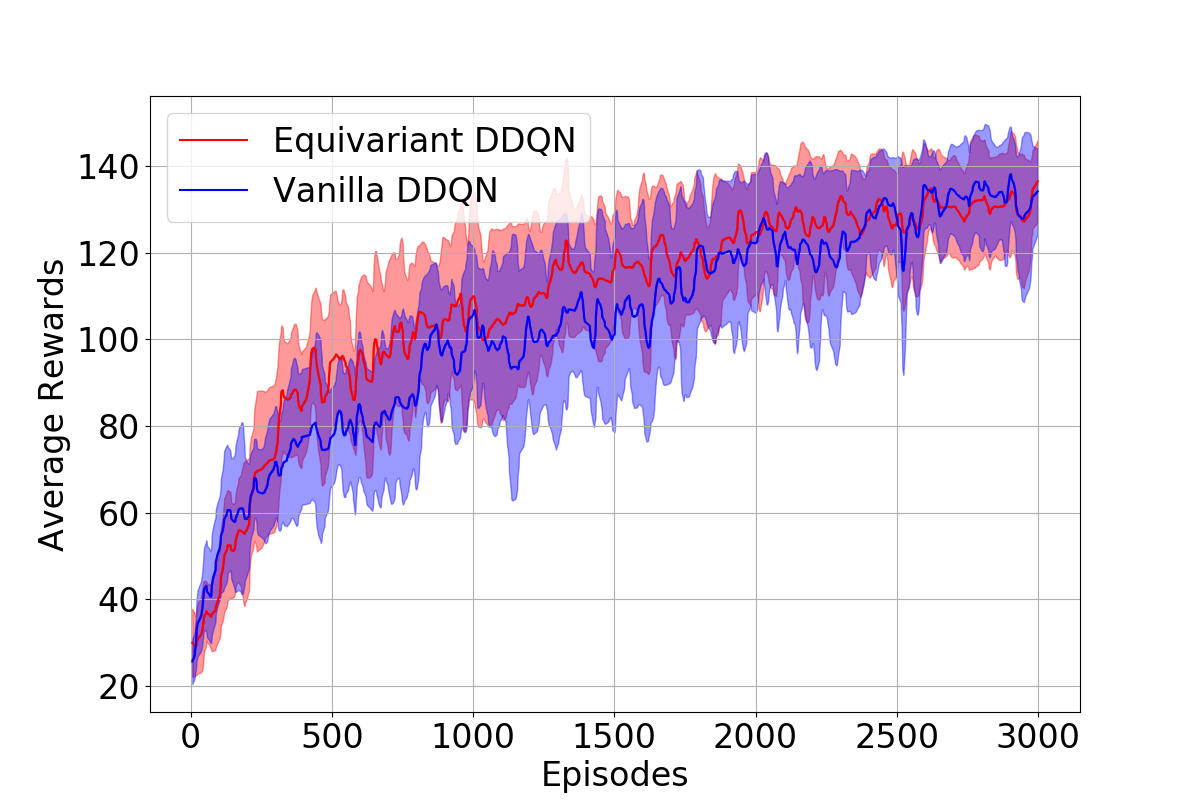}
		\captionsetup{labelformat=empty}
		  \caption*{Pacman}
        \label{fig:pacman}
        \end{minipage}
	\caption{Plots of evolution of average rewards with the number of episodes. We also show the confidence intervals over 10 different seeds. Note that the plots are smoothed with a 1D Gaussian filter with $\sigma$=3 for better visualization.}
\label{fig:results}
\vspace{-0.7cm}
\end{figure}
We first consider the performance of a carefully designed equivariant DDQN, keeping in mind the symmetry of the game (refer to Appendix \ref{netarc}), compared to a vanilla DDQN. For a fair comparison, we keep the settings of the environment and hyperparameters the same for all the experiments\footnote{Link to the code: \href{https://github.com/arnab39/EquivariantDQN}{https://github.com/arnab39/EquivariantDQN} }. 
We report in Figure \ref{fig:results} the evolution of rewards collected over the training episodes for both the models in the Snake and the Pacman environments. Our proposed model attains a $30\%$ improvement in average reward collected after training for $3000$ episodes in the highly symmetric Snake environment. It also learns faster with a $90\%$ reduction in the number of parameters. This verifies our hypothesis that parameters required to learn policies of the identity transformation would be sufficient to generalize to optimal policies in other transformations for the Snake environment. In the case of Pacman, we notice our model performs slightly better in the initial episodes, with a $34\%$ reduction in the number of parameters. But once both the models have seen enough samples, the margin of difference vanishes. In Appendix \ref{add}, we show that the proposed method gives similar results with other subsequent improvements, such as using DDQN with priority replay and the Dueling architecture.
\vspace{-0.1cm}

We further investigate the usefulness of the inherent inductive bias in the model in transfer learning with respect to the affine transformation of the environment screen. For this part, we remove the group restriction from Equivariant DDQN of the Pacman game and make the feature extractor $D_4$ equivariant. First we train both the Vanilla and Equivariant model. We then change the environment by rotating the input screen by 90 degrees clockwise ($r$). Leaving the rest of the network frozen, we retrain the final linear layer for this new environment. We show in Table \ref{tab_rot} that while a regular CNN based feature extractor fails, the Equivariant feature extractor can still find a decent policy after learning the linear layers for certain epochs. The results of a simple path planning problem like Snake, indicate that our model can, in principle, be extended to more complex continuous path planning problems such as in UAVs \cite{zhang2015geometric, challita2018deep}. Such scenarios would benefit both from faster learning due to increased sample efficiency and viewpoint transformation equivariant features for optimal policy learning, which can generalize to new transformations of the environment.  

\begin{table}[!ht]
    \centering
    \begin{tabular}{ccc}
    \cline{1-3}
       \makecell{Transformation}  & Vanilla DDQN & Equivariant DDQN \\
       \cline{1-3}
       \makecell{$e$}  & 129 $\pm$ 2.3  &    125 $\pm$ 4.5 \\ 
\makecell{$r$}  & 48.9 $\pm$ 2   &    99 $\pm$ 4.7     \\ 
\makecell{$r^2$} & 53 $\pm$ 3.5  &   104 $\pm$ 3.4   \\
\makecell{$r^3$} & 51 $\pm$ 1.9  &   98 $\pm$ 3.9  \\ 
 \cline{1-3}
    \end{tabular}
    \caption{Average reward over 200 episodes of Pacman for 5 seeds reported with a confidence level of 95\% for different environment transformations. $e$ is the original screen.}
    \label{tab_rot}
    \vspace{-0.5cm}
\end{table}

\section{Conclusions and future work}
\label{conclusion}

We have introduced an Equivariant Deep-Q learning algorithm, and have demonstrated that it provides a considerable boost to performance with parameter and sample efficiency when carefully designed for highly symmetric environments. We have also shown that this approach generalizes policies well to new unseen environments obtained by an affine transformation of the original environment. Although invariant models in supervised learning were shown to make the models robust, to the best of our knowledge, this is the first time equivariant learning has been proposed in a Deep RL framework. In follow-up work, we plan to implement continuous rotation and reflection group equivariance using an irreducible representation of the $O_2$ group, for more challenging path planning environments, with an extension to a continuous action space. 

\section{Acknowledgement}
We would like to thank Gabriele Cesa for his valuable comments on the E(2)-equivariant convolution.

\nocite{langley00}

\bibliography{example_paper}
\bibliographystyle{icml2019}

\appendix
\begin{figure*}[!ht]
    \centering
    \includegraphics[width=0.7\textwidth,height=5.5cm]{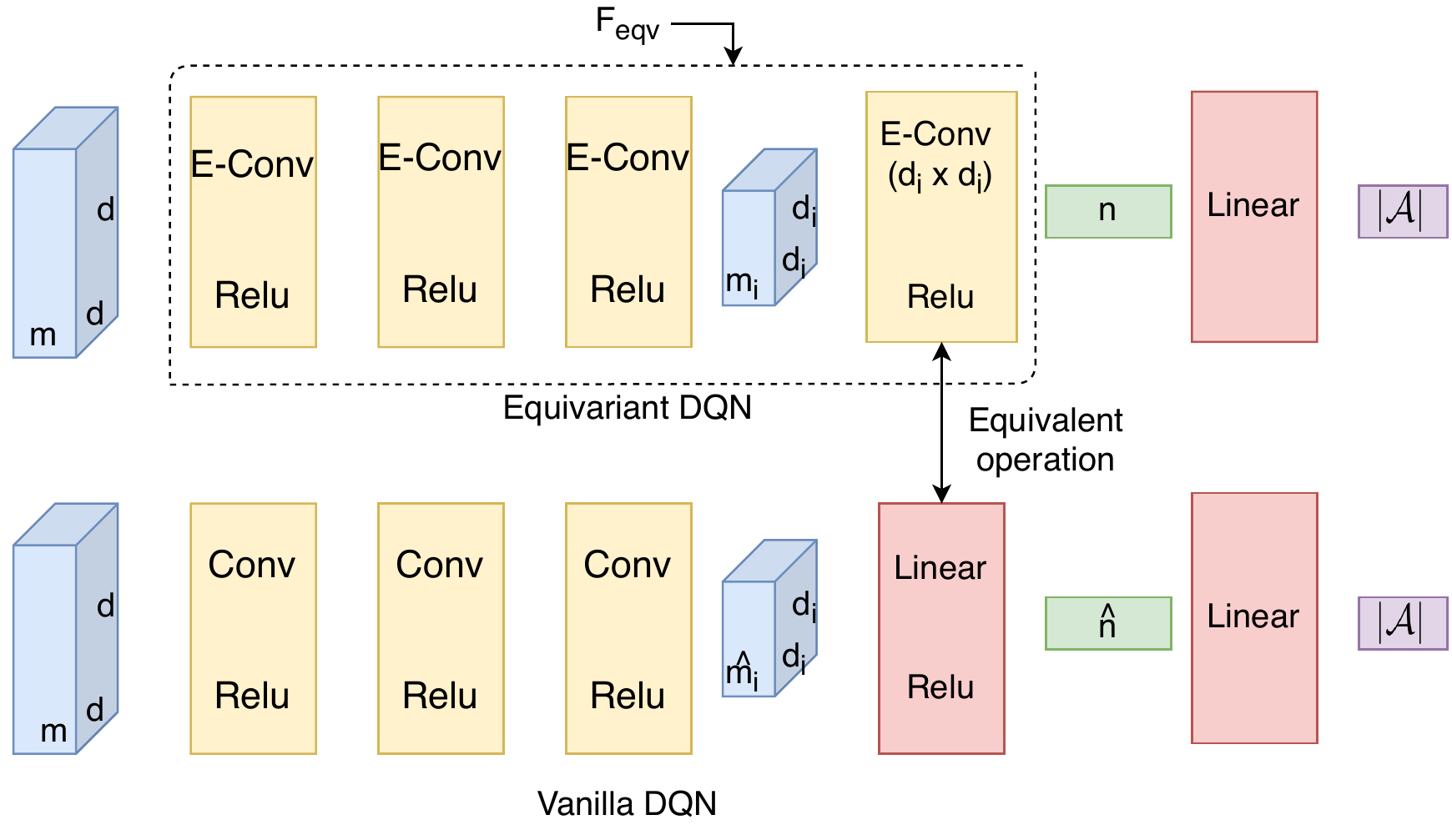}
    \caption{Juxtaposed Network architecture}
    \label{fig:network}
    \label{arch}
\end{figure*}

\section{Group representation and restriction in feature fields}
\label{grprep}
We now discuss how one would choose a group ($g$) and its representation ($\rho$) to define a feature field. The group's choice mainly depends on the problem we are tackling and to which kinds of transformation we wish the network to output equivariantly. We have several options for E(2)-equivariant convolution, starting from discrete rotations and reflections ($D_N$) to continuous rotation and reflection ($O(2)$). Once a group is chosen, we need to choose its representation. The most common ones are trivial, irreducible, regular and quotient representations. The representation chosen determines the dimension $c$ of a feature vector. While a trivial representation implies scalar features with dimension $1$ the regular representation uses an $N$-dimensional feature field, where $N$ denotes the order of the group we are using. Even though a regular representation was shown to perform the best\cite{weiler2019general}, it is computationally infeasible to use it when using higher-order groups. In such a case, we use an irreducible representation, which takes the smallest dimension while leaving the representation of all the group elements unique. 

Let us assume that we are working with a generic $D_N$ group with its regular representation. The next thing we need to choose is the number of feature fields for each intermediate layer. Together the chosen representation and number of feature fields contribute to the dimension of the stack $\bigoplus_i f_i$ of intermediate feature fields, which further determines the depth of the convolution kernel we are using between two feature fields. Although increasing the number of feature fields increases the network's capacity, this comes at the cost of increased computation during a single forward pass. 

In an environment where we have a global $D_N$ symmetry, where we want $D_N$ equivariant features and $N>1$, we can directly choose the group and keep it throughout. But in most environments where there is usually a global $D_1$ symmetry and occasionally local $D_N$ symmetry, using the same representation throughout would be futile as this is accompanied by order of $N$ increase in feature field dimension. To alleviate this problem, we start with a higher-order group $D_N$ where $N>1$ and as we go deeper into our $Q$ network, we restrict it to its subgroups($\leq D_N$). This makes the network more computationally efficient while still extracting an equivariant feature vector.    
 \vspace{-0.2cm}
\section{Network Architecture}
\label{netarc}
The baseline Vanilla DDQN used in this work is similar to the one used in \cite{mnih2013playing}, which has an output dimension equal to the number of actions. As shown in Figure \ref{arch}, our proposed Equivariant DDQN architecture mainly replaces the Vanilla convolutions($Conv$) and the second last linear layer with equivariant convolutions($E-Conv$). We call this an equivariant feature extractor. We want to emphasize on the last E-Conv layer and point-out that its operation is similar to the second last linear layer in a Vanilla DDQN. As we use the filter size of the dimension of feature size before that layer, all the information is captured as a weighted sum into a 1-D vector. Although this is the same as the flattening of the feature and then applying a linear layer, using $E-Conv$ renders the output vector equivariant. 

The final linear layer is the same for both and maps them to $Q$-values for each action. Notice, as mentioned in Appendix \ref{grprep}, the group representation and the number of feature field will determine the sizes of intermediate features. We aim to make both networks similar with respect to computation time while not comprising the capacity of the Equivariant model. Below we provide the architecture of the Equivariant and Vanilla DDQN for both Snake and Pacman. We denote a basic convolution by: 
$Conv(filtersize\times filtersize, inchannels, outchannels, stride, padding)$ and equivariant one by: $E-Conv(filtersize\times filtersize, infields, outfields, stride, padding)[Group]$ The group restriction operation is denoted by: $GrpRes[Group\to Subgroup]$. In the Vanilla and Equivariant DDQN, we denote the size of the output of the third convolution by $\hat{m}_i\times d_i\times d_i$ and $m_i\times d_i\times d_i$ respectively. Using this, we give the exact architecture of both the networks below.
\vspace{-0.3cm}
\subsection{Snake}
\subsubsection{Vanilla DDQN}
$Conv(7\times7, m, 32, 2, 2) - ReLU\\ 
Conv(5\times5, 32, 64, 2, 1) - ReLU\\
Conv(5\times5, 64, 64, 1, 1)- ReLU\\
Linear(\hat{m}_i\times d_i\times d_i, 256)-ReLU\\
Linear(256, |\mathcal{A}|)$
\subsubsection{Equivariant DDQN}
$E-Conv(7\times7, m, 8, 2, 2)[D_4] - ReLU\\ 
E-Conv(5\times5, 8, 12, 2, 1)[D_4] - ReLU\\
E-Conv(5\times5, 12, 12, 1, 1)[D_4]- ReLU\\
E-Conv(d_i\times d_i, 12, 32, 1, 0)[D_4]-ReLU\\
Linear(256, |\mathcal{A}|)$

\subsection{Pacman}
\subsubsection{Vanilla DDQN}
$Conv(7\times7, m, 32, 4, 2) - ReLU\\ 
Conv(5\times5, 32, 64, 2, 2) - ReLU\\
Conv(5\times5, 64, 64, 2, 1)- ReLU\\
Linear(\hat{m}_i\times d_i\times d_i, 512)-ReLU\\
Linear(512, |\mathcal{A}|)$
\subsubsection{Equivariant DDQN}
$E-Conv(7\times7, m, 8, 2, 2)[D_4] - ReLU\\ 
E-Conv(5\times5, 8, 16, 2, 1)[D_4] - ReLU\\
GrpRes[D_4\to D_1]\\
E-Conv(5\times5, 16, 64, 1, 1)[D_1]- ReLU\\
E-Conv(d_i\times d_i, 64, 384, 1, 0)[D_1]-ReLU\\
Linear(768, |\mathcal{A}|)$

\begin{table}[!ht]
    \centering
    \begin{tabular}{ccc}
    \cline{1-3}
       \makecell{Network Type}  & Vanilla DQN & \makecell{Number of\\ Parameters} \\
       \cline{1-3}
       \makecell{Vanilla DDQN}  & Snake &    583.46k  \\ 
\makecell{Equivariant DDQN}  & Snake &    57.7k     \\ 
\makecell{Vanilla DDQN} & Pacman &   984.36k    \\
\makecell{Equivariant DDQN} & Pacman &   649.93k    \\ 
 \cline{1-3}
    \end{tabular}
    \caption{This table gives the number of parameters of both the Networks.}
    \label{tab_complexity}
\end{table}
Although there is a difference in the number of channels and feature fields, the overall runtime of the DDQN algorithms with both the networks are similar. The forward pass of the Equivariant network is more computationally expensive as the total dimension of the stack of feature fields in some layers is more than the number of channels in the Vanilla network. But this is partially compensated for during the backpropagation where we are updating fewer parameters in an Equivariant network. Note that, in general, adding feature fields increases the capacity at the cost of computation, but we keep the total cost with respect to the Vanilla model in mind while choosing them. Also, as the Pacman environment is globally symmetric to the $D_1$ group, we restrict the group once $D_4$ symmetric lower level features are extracted, which also reduces the dimension of representation and hence the computation cost significantly. It is interesting to note that higher-order symmetry in the environment leads to fewer parameters than the Vanilla DQN. 
\vspace{-0.2cm}
\section{Additional Results}
\label{add}
In this section, we provide some additional results of our proposed method applied to  DDQN with priority replay and Dueling DDQN in the Snake Environment. We show that our proposed models outperform the Vanilla models in both the cases, which demonstrates that our approach scales to handle different algorithms. Note that we used a lower learning rate for Dueling DQN to stabilize training.   

 \begin{figure}[!ht]
		\begin{minipage}{\columnwidth}
		\centering
        \includegraphics[width=\columnwidth,height=4.3cm]{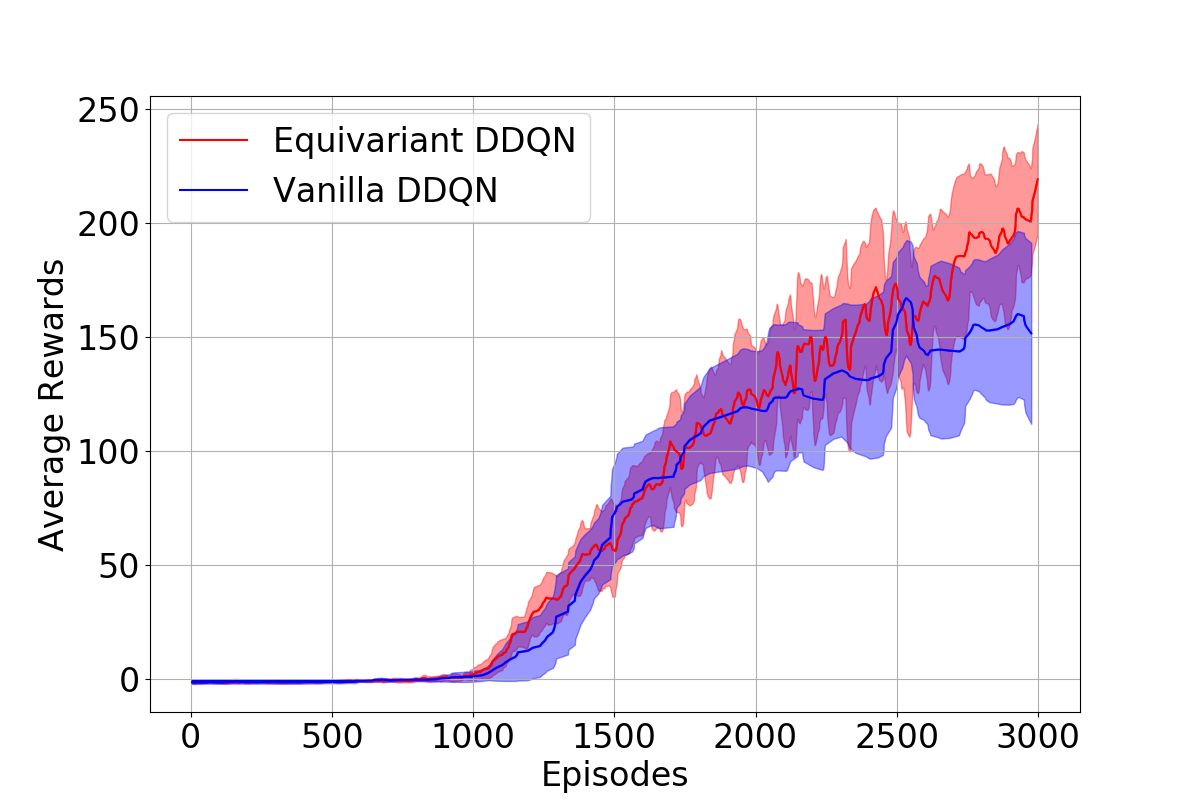}
		\captionsetup{labelformat=empty}
		\caption*{DDQN with priority replay in Snake}
 		\label{fig:priority_snake}
		\end{minipage}
		\begin{minipage}{\columnwidth}
		\centering
		\includegraphics[width=\columnwidth, height=4.3cm]{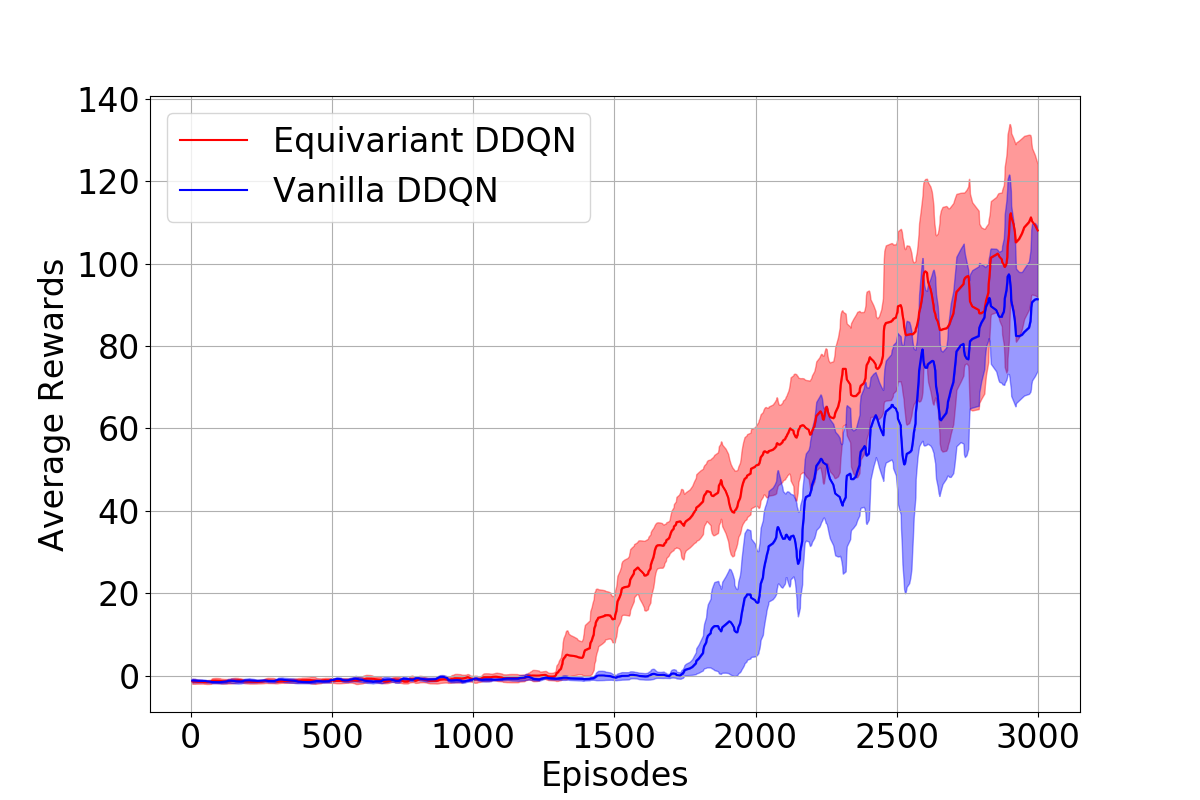}
		\captionsetup{labelformat=empty}
	    \caption*{Dueling DQN in Snake}
        \label{fig:dueling_snake}
        \end{minipage}

\caption{Plots of evolution of average rewards with the number of episodes over 10 different seeds and are smoothed with a 1D Gaussian filter with $\sigma$=3 for better visualization.}
\label{fig:results_appendix}
\end{figure}




\end{document}